\documentclass[10pt,twocolumn,letterpaper]{article}

\usepackage[pagenumbers]{cvpr} 

\definecolor{cvprblue}{rgb}{0.21,0.49,0.74}
\usepackage[pagebackref,breaklinks,colorlinks,allcolors=cvprblue]{hyperref}
\usepackage{xcolor}
\usepackage{algorithm}
\usepackage{algpseudocode}
\usepackage{array}
\usepackage{graphicx}
\usepackage{amssymb}
\usepackage{pifont}

\def\paperID{} 
\def\confName{CVPR}
\def\confYear{2026}

\title{TINA: Text-Free Inversion Attack for Unlearned Text-to-Image Diffusion Models}

\author{
Qianlong Xiang$^{1,2,3}$, Miao Zhang$^{1,}$\footnotemark[2], Haoyu Zhang$^{1,4}$, Kun Wang$^{5}$, Junhui Hou$^{2}$, Liqiang Nie$^{1,3,}$\footnotemark[2]\\
$^1$Harbin Institute of Technology (Shenzhen)
$^2$City University of Hong Kong
$^3$Shenzhen Loop Area Institute\\
$^4$Peng Cheng Laboratory
$^5$Shandong University\\
\texttt{\normalsize{\url{https://github.com/qianlong0502/TINA}}}
}

\begin{document}
\maketitle

\renewcommand{\thefootnote}{\fnsymbol{footnote}}
\footnotetext[2]{Corresponding authors}
\newcolumntype{P}[1]{>{\centering\arraybackslash}p{#1}}
\begin{abstract}
Although text-to-image diffusion models exhibit remarkable generative power, concept erasure techniques are essential for their safe deployment to prevent the creation of harmful content.
This has fostered a dynamic interplay between the development of erasure defenses and the adversarial probes designed to bypass them, and this co-evolution has progressively enhanced the efficacy of erasure methods.
However, this adversarial co-evolution has converged on a narrow, text-centric paradigm that equates erasure with severing the text-to-image mapping, ignoring that the underlying visual knowledge related to undesired concepts still persist.
To substantiate this claim, we investigate from a visual perspective, leveraging DDIM inversion to probe whether a generative pathway for the erased concept can still be found.
However, identifying such a visual generative pathway is challenging because standard text-guided DDIM inversion is actively resisted by text-centric defenses within the erased model.
To address this, we introduce TINA, a novel Text-free INversion Attack, which enforces this visual-only probe by operating under a null-text condition, thereby avoiding existing text-centric defenses.
Moreover, TINA integrates an optimization procedure to overcome the accumulating approximation errors that arise when standard inversion operates without its usual textual guidance.
Our experiments demonstrate that TINA regenerates erased concepts from models treated with state-of-the-art unlearning.
The success of TINA proves that current methods merely obscure concepts, highlighting an urgent need for paradigms that operate directly on internal visual knowledge.
\end{abstract}

\begin{figure}
    \centering
    \includegraphics[width=\linewidth]{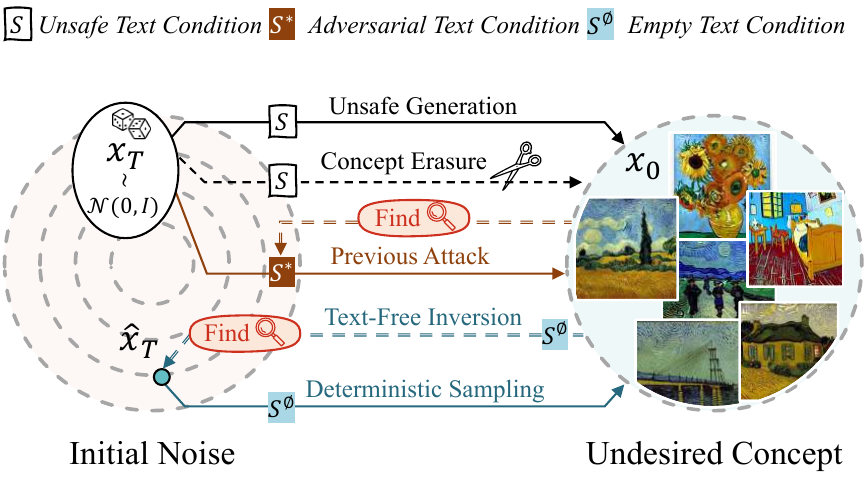}
    \caption{
    Conceptual overview of text-centric erasure vulnerabilities and our TINA attack.
    \textbf{Concept Erasure} usually severs the link between a specific text condition and the undesired concept.
    \textcolor[RGB]{142, 65, 12}{\textbf{Previous Attacks}} remain text-centric, finding adversarial text condition to reactivate the concept.
    \textcolor[RGB]{38, 104, 125}{\textbf{Our TINA}} bypasses the text pathway entirely. Using an empty text condition, it finds a noise to regenerate the concept, proving the visual knowledge persists in the existing erased models.
    }
    \label{fig:intro}
\end{figure}

\section{Introduction}
\label{sec:intro}
With the rapid development of deep learning \cite{vaswani2017attention,zhang2023attribute,zhang2024multi,zhang2025spatial,li2026optimus3,li2026optimusvla,wang2025unifying,wang2025decad,liu2025understanding,wang2026cross,wang2024explicit,wang2025redundancy}, generative artificial intelligence, particularly Text-to-Image (T2I) diffusion models such as Stable Diffusion is fundamentally reshaping the landscape of digital content creation~\cite{rombach2022high,nichol2021glide,ramesh2021zero,saharia2022photorealistic,xiang2025dkdm}.
Their remarkable capacity for creative synthesis has catalyzed countless applications, from art and design to entertainment and media.
However, since these models are typically trained on massive, unfiltered datasets scraped from the Internet~\cite{schuhmann2022laion}, their proliferation brings a turbulent undercurrent of ethical and safety challenges.
The same models that generate breathtaking art can be misused to create deepfakes that violate personal privacy~\cite{carlini2023extracting}, imitate artistic styles that infringe on copyright~\cite{jiang2023ai,roose2022ai,somepalli2023diffusion}, and produce harmful Not-Safe-For-Work (NSFW) imagery~\cite{zhang2024generate,hunter2023ai,schramowski2023safe}, posing significant risks to intellectual property and societal well-being.

To address these challenges, the field has converged on the paradigm of \textit{Concept Erasure}, a specialized form of \textit{Machine Unlearning}~\cite{xu2023machine}.
This paradigm aims to mitigate risks by directly modifying model parameters.
For instance, approaches such as Erasing Stable Diffusion (ESD) and Ablating Concepts (AC) retrain specific model layers to remap a forbidden concept representation to that of a neutral or null concept, thereby severing the explicit link between a text prompt and its corresponding visual output~\cite{gandikota2023erasing,kumari2023ablating}.
Other methods like Forget-Me-Not (FMN) minimize the attention maps corresponding to the target concept~\cite{zhang2024forget}, while Unified Concept Editing (UCE) applies efficient, closed-form edits to the cross-attention weights~\cite{gandikota2024unified}.
Despite their distinct mechanisms, these approaches share a common, text-centric operational principle, equating erasure with merely severing its \textit{text-image mapping}.

The text-centric paradigm, as conceptualized in Figure~\ref{fig:intro}, has consequently guided adversarial attacks almost exclusively to the textual domain.
Initial red-teaming efforts involve prompt-level manipulations, attempting to find alternative phrasings or complex prompts that reactivate the target concept~\cite{chin2023prompting4debugging,zhang2024generate,tsai2023ring}.
Moreover, the more sophisticated embedding-space attacks, which move beyond discrete tokens, remain fundamentally dependent on the textual conditioning pathway~\cite{pham2023circumventing}.
Generally, these red-teaming methods aim to discover new token embeddings~\cite{pham2023circumventing} or optimize an adversarial text prompt~\cite{zhang2024generate} that serves as a proxy for the erased concept.
In response to this threat, an interplay between text-based attacks and defenses has driven the subsequent development of more resilient safeguards, such as AdvUnlearn~\cite{zhang2024defensive} and STEREO~\cite{srivatsan2025stereo}.
This adversarial co-evolution has produced defenses that demonstrate increasing efficacy, thereby fostering an appearance of robust model security.

However, in this work, we contend that this competitive co-evolution of text-centric methods rests on a fundamental and fatal assumption.
Specifically, existing paradigm mistakenly equates the erasure of a \textit{text-to-image link} with the far more complex task of eliminating the underlying \textit{visual knowledge} from the parameter space.
Consequently, existing methods critically ignore the fact that even if textual inputs can no longer induce the target concept, the corresponding visual knowledge may still persist, untapped, within the model.
To challenge this dominant paradigm and substantiate our claim, we introduce a core hypothesis:

\textit{A deterministic generative pathway for an erased concept persists within the model, even after its text-image mappings are removed, allowing for its regeneration in a completely text-free manner.}

To validate this hypothesis from a purely visual perspective, thereby challenging the efficacy of these text-centric defenses, we introduce TINA, a Text-free INversion Attack.
Unlike prior embedding-space attacks that seek a new textual proxy for the erased concept~\cite{pham2023circumventing}, TINA is a novel framework designed to completely bypass the textual conditioning pathway.
To achieve this, TINA leverages Denoising Diffusion Implicit Models (DDIM)~\cite{song2020denoising} inversion as a visual probe (illustrated in Figure~\ref{fig:intro}), allowing us to investigate the visual knowledge embedded within the model directly.
However, identifying such a generative pathway is challenging because standard DDIM inversion relies heavily on textual guidance.
Operating under the null-text condition necessary to evade defenses consequently introduces significant imprecision.
To address this imprecision, TINA incorporates a novel optimization procedure that corrects for the compounding errors of standard inversion, allowing it to precisely map a target image back to a latent noise vector.
Serving as the specific initial noise for the deterministic DDIM sampling process, this resulting latent vector can lead an erased model to regenerate the forbidden content through a deterministic generative trajectory, entirely bypassing the text-conditioning mechanism.
The success of this text-free reconstruction provides compelling evidence that current erasure methods merely obscure concepts by severing text-image links, rather than truly expunging the underlying visual knowledge embedded deep within the model parameters.
These findings serve as a critical alert for the AI safety community, highlighting the urgent need for more robust unlearning paradigms that operate directly on the internal visual representations within T2I models.

To sum up, our contributions are as follows.

\begin{itemize}
    \item To our knowledge, we are the first to identify the fundamental, text-centric vulnerability in the current paradigm of concept erasure for T2I diffusion models.
    \item We propose TINA, a new text-free attack paradigm that employs an optimization-based inversion to identify the underlying generative trajectory for a supposedly erased concept, from a purely visual perspective.
    \item Experiments demonstrate that our TINA can rediscover erased concepts from models unlearned by state-of-the-art methods, thereby exposing a fundamental flaw needing urgent attention.
\end{itemize}

\section{Related Work}
\label{sec:related}

\subsection{Text-to-Image Diffusion Models}
The advent of Text-to-Image (T2I) diffusion models, such as Stable Diffusion~\cite{rombach2022high}, DALL-E 2~\cite{ramesh2022hierarchical}, and Imagen~\cite{saharia2022photorealistic}, has marked a paradigm shift in digital content generation. These models operate through a two-stage process: a forward diffusion process that incrementally adds noise to an image, and a learned reverse process that denoises a random vector back into a coherent image~\cite{ho2020denoising}. To achieve computational efficiency, models like Stable Diffusion perform this process in a lower-dimensional latent space~\cite{rombach2022high}. Critically, their generative process is guided by textual prompts, typically encoded and injected into the denoising network via a cross-attention mechanism~\cite{vaswani2017attention}. This tight coupling between text and image synthesis is the source of their remarkable controllability, but also the root of significant safety and ethical concerns, as the models can be prompted to produce NSFW imagery~\cite{zhang2024generate,schramowski2023safe}, imitate copyrighted styles~\cite{jiang2023ai,somepalli2023diffusion}, or violate personal privacy~\cite{carlini2023extracting}.

\subsection{Concept Erasure in Diffusion Models}
To address the misuse of T2I models, concept erasure has emerged as a critical field.
While methods like SLD~\cite{schramowski2023safe} focus on inference-time interventions, the dominant paradigm involves fine-tuning-based methods that modify model parameters.
This paradigm originated with foundational works like ESD~\cite{gandikota2023erasing} and AC~\cite{kumari2023ablating}, which retrained models to break associations with specific text prompts.
Subsequently, this was followed by more efficient and targeted techniques, such as FMN~\cite{zhang2024forget}, UCE~\cite{gandikota2024unified}, MACE~\cite{lu2024mace}, and the adapter-based SPM~\cite{lyu2024one}, often focusing on cross-attention layers.
Concurrently, general unlearning frameworks like SalUn~\cite{fan2023salun}, Scissorhands~\cite{wu2024scissorhands}, and EraseDiff~\cite{wu2024erasediff} were proposed for the concept erasure task.
As text-based attacks began to bypass these initial erasure methods, advanced adversarially robust methods were developed, including RECE~\cite{gong2024reliable}, AdvUnlearn~\cite{zhang2024defensive}, and STEREO~\cite{srivatsan2025stereo}.
These methods usually integrate attack considerations into their design to defend against adversarial prompts and embeddings.
Ultimately, this co-evolution of text-based attacks and defenses has solidified a dominant, text-centric paradigm: most of them operate by identifying or severing the \textit{text-to-image mapping}, rather than addressing the underlying visual knowledge.

\subsection{Adversarial Attacks on Concept Erasure}
The prevailing text-centric assumption of erasure methods has consequently shaped the landscape of adversarial attacks, which primarily seek to circumvent these textual defenses.
These attacks can be classified by their operational domain.
The most straightforward approaches operate in the discrete token space, employing circumvention strategies to find alternative phrases that trigger the erased concept, such as P4D~\cite{chin2023prompting4debugging}, RAB~\cite{tsai2023ring}, MMA~\cite{yang2024mma}, and UDA~\cite{zhang2024generate}.
A more sophisticated class of attacks operates in the continuous embedding space.
These methods aim to find a novel token embedding that acts as a proxy for the forbidden concept, effectively creating a new activation pathway to the visual knowledge.
For instance, CCE~\cite{pham2023circumventing} leverages textual inversion to discover surrogate embeddings for erased concepts.

Critically, despite their varying levels of sophistication, from simple phrasings to optimized vectors, these existing attacks share a fundamental limitation: most of them presuppose that the generative capabilities of the model must be accessed via its text-conditioning pathway.
They focus on finding a new textual or embedding-space vector to bypass modified controls.
In stark contrast, our work, TINA, challenges this entire paradigm.
We pivot to a fundamentally different attack surface by postulating that visual knowledge persists independently of any textual control.
Accordingly, we propose an attack that circumvents the text-conditioning mechanism entirely, identifying and exploiting a deterministic, non-textual generative trajectory to regenerate forbidden content without textual guidance.

\section{Method}
\label{sec:method}

\subsection{Preliminaries}
\label{sec:preliminaries}

\paragraph{Latent Diffusion Models.}
Latent Diffusion Models (LDMs)~\cite{rombach2022high} generate images by reversing a $T$-step diffusion process in a compressed latent space. An image $x$ is first encoded into its latent $z_0 = \mathcal{E}(x)$ by a VAE encoder. The forward process creates a sequence of noised latents $\{z_1, \dots, z_T\}$.
A denoising U-Net $\epsilon_{\theta}$ learns to reverse this process. At each timestep $t$, it predicts the noise $\epsilon$ in $z_t$, guided by a textual condition $c$. This condition is produced by a text encoder $\mathcal{T}_{\text{text}}$ from a prompt $y$ and integrated via cross-attention. The training objective minimizes the noise prediction error:
$$
\mathcal{L} = \mathbb{E}_{z_0, y, \epsilon, t} \left[ \left\| \epsilon - \epsilon_{\theta}(z_t, t, c) \right\|_2^2 \right].
$$
This objective trains $\epsilon_{\theta}$ to denoise $z_t$ back towards $z_0$ in a manner faithful to the condition $c$. The final image is retrieved using the VAE decoder $\mathcal{D}$.

\begin{figure*}[t]
    \centering
    \includegraphics[width=\linewidth]{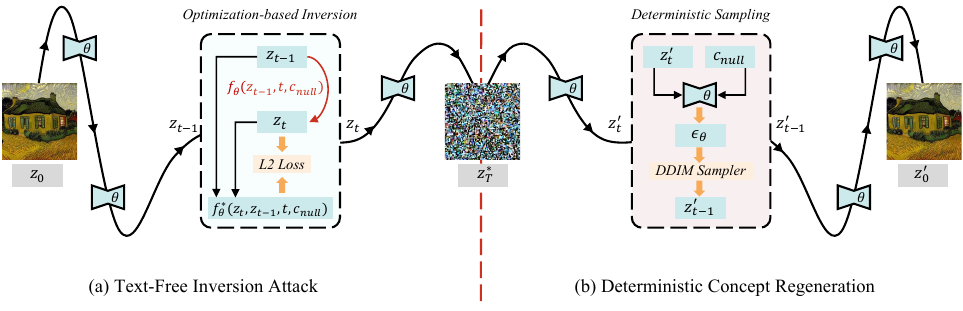}
    \caption{The TINA (Text-free INversion Attack) framework. (a) \textbf{Text-Free Inversion Attack:} An optimization-based, null-text ($c_{\text{null}}$) inversion finds the unique initial noise $z_T^*$ corresponding to a target image $z_0$. This optimization corrects the errors from standard inversion. (b) \textbf{Deterministic Concept Regeneration:} The \textit{same} sanitized model $\theta$ uses $z_T^*$ and $c_{\text{null}}$ to deterministically regenerate the target concept $z_0'$, proving the visual knowledge persists despite erasure.}
    \label{fig:overview}
\end{figure*}

\paragraph{DDIM Sampling.}
Denoising Diffusion Implicit Models (DDIM)~\cite{song2020denoising} enable deterministic generative sampling from LDMs. This is achieved by formulating the reverse process as a non-Markovian, discretized Probability Flow Ordinary Differential Equation (PF-ODE) with zero variance. Consequently, for a fixed model $\theta$ and condition $c$, any initial noise $z_T \sim \mathcal{N}(0, \mathbf{I})$ maps to a unique generated latent $z_0$.

This deterministic path is defined by an iterative update rule. Given the latent $z_t$, the model first predicts the clean latent $\hat{z}_0(z_t)$:
\begin{equation}
\label{eq:ddim_pred_z0}
\hat{z}_0(z_t) = \frac{z_t - \sqrt{1-\alpha_t}\epsilon_\theta(z_t, t, c)}{\sqrt{\alpha_t}}.
\end{equation}
The preceding latent $z_{t-1}$ is then computed as:
\begin{equation}
\label{eq:ddim_sample}
z_{t-1} = \sqrt{\alpha_{t-1}} \hat{z}_0(z_t) + \sqrt{1-\alpha_{t-1}} \cdot \epsilon_\theta(z_t, t, c),
\end{equation}
where $\{\alpha_t\}_{t=1}^T$ are the fixed noise schedule parameters.

\paragraph{DDIM Inversion.}
Building upon the deterministic nature of DDIM sampling, DDIM inversion provides a mechanism to reverse the generative process. Its objective is to map a given image, represented by its initial latent $z_0$, back to the unique starting noise vector $z_T$ from which it can be deterministically generated. This capability is crucial for applications like real image editing and analysis.

Ideally, the inversion would be a perfect algebraic reversal of the sampling step in Eq.~(\ref{eq:ddim_sample}), which can be expressed as:
\begin{equation}
\label{eq:ddim_inversion_ideal}
z_t = C_1(t) z_{t-1} + C_2(t) \cdot \epsilon_\theta(z_t, t, c),
\end{equation}
where 
\begin{gather}
C_1(t) = \frac{\sqrt{\alpha_t}}{\sqrt{\alpha_{t-1}}}, \label{eq:c1_coeff} \\
C_2(t) = \sqrt{1-\alpha_t} - \sqrt{\frac{\alpha_t(1-\alpha_{t-1})}{\alpha_{t-1}}}. \label{eq:c2_coeff}
\end{gather}
However, a practical challenge emerges: computing the latent $z_t$ requires a noise prediction $\epsilon_\theta(z_t, t, c)$, which itself depends on $z_t$, as shown in Eq.~(\ref{eq:ddim_inversion_ideal}). To break this circular dependency, standard DDIM inversion employs a key approximation: it estimates the required noise using the latent from the previous step, $z_{t-1}$, at timestep $t-1$. This results in the following iterative formula to approximate $z_t$ from $z_{t-1}$:
\begin{equation}
\label{eq:ddim_inversion_approx}
z_t \approx f_{\theta}(z_{t-1}, t, c),
\end{equation}
where
\begin{equation}
\label{eq:ddim_inversion_approx_f}
f_{\theta}(z_{t-1}, t, c)=
C_1(t) z_{t-1} + C_2(t) \cdot \epsilon_\theta(z_{t-1}, t-1, c).
\end{equation}
Starting from the clean image latent $z_0$, this equation is applied sequentially for $t=1, \dots, T$ to trace the trajectory back to an approximate initial noise $\hat{z}_T$. This vector serves as an estimate of the true starting point in the noise space required for reconstructing the original image.

\subsection{TINA: Text-free INversion Attack}
\label{sec:tina}

In this section, we introduce our TINA framework, starting with the attack pipeline (Section~\ref{subsec:pipeline}), then analyzing why the naive standard inversion fails to meet the objective in the pipeline (Section~\ref{subsec:challenge}), and finally detailing our optimization-based solution (Section~\ref{subsec:solution}).

\subsubsection{The TINA Attack Pipeline}
\label{subsec:pipeline}

The central hypothesis of TINA is that even within a model subjected to concept erasure, deterministic generative trajectories persist that are entirely independent of textual conditioning.
Our framework is designed to expose and exploit this vulnerability through a two-phase attack pipeline, as illustrated in Figure~\ref{fig:overview}.

The process begins with a concept-erased model $\epsilon_{\theta}$ and a target image $x$ representing the erased concept.
As illustrated in Figure~\ref{fig:overview}, the attack first performs a \textbf{Text-Free Inversion} to find the precise initial noise vector $z_T^*$ that corresponds to the target image latent $z_0$.
This inversion phase uses only the sanitized model $\epsilon_{\theta}$ and a null-text condition $c_{\text{null}}$.
Following this, in the \textbf{Deterministic Regeneration} phase, this optimized $z_T^*$ is fed back into the \textit{same} erased model $\epsilon_{\theta}$.
A standard deterministic DDIM sampling process is then executed, crucially starting from this specific $z_T^*$ and operating entirely under the null-text condition $c_{\text{null}}$.

The successful reconstruction of the target concept $x'$ serves as definitive proof that the visual knowledge persists, independent of textual controls.
The deterministic regeneration step is a standard DDIM sampling procedure; the core technical challenge lies in the initial inversion step: accurately identifying the starting noise vector $z_T^*$ for a target image $x$ representing the erased concept.

\begin{figure}[t]
    \centering
    \includegraphics[width=0.92\linewidth]{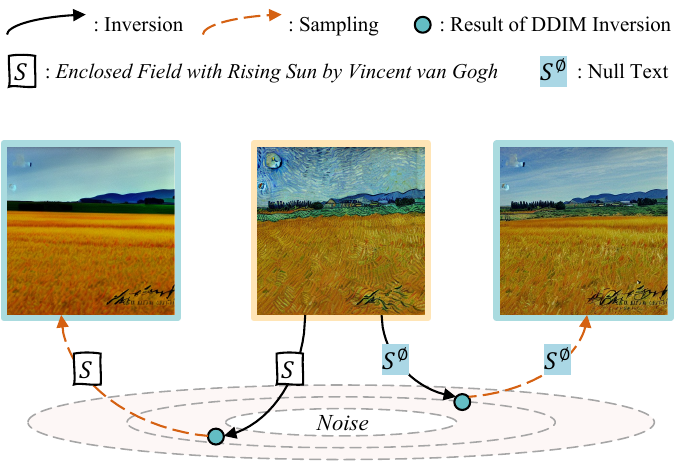}
    \caption{Standard DDIM inversion fails to find the generative trajectory. The text-guided path ($S$) is blocked by the erasure, while the null-text path ($S^\emptyset$) drifts due to approximation errors, failing to restore the target concept.}
    \label{fig:ddim_inv}
\end{figure}

\subsubsection{Limitations of Standard Inversion}
\label{subsec:challenge}

A naive approach to finding $z_T^*$ would be to apply the standard DDIM inversion process to the target image latent $z_0$.
However, this method is fundamentally flawed and fails in two distinct ways, as illustrated in Figure~\ref{fig:ddim_inv}.
First, if the inversion is attempted using the text condition $c$ of the target concept (e.g., ``Van Gogh style''), the sanitized model $\epsilon_{\theta}$ actively counteracts this prompt, leading to a complete failure in reconstruction.
This confirms that text-centric defenses are indeed working against text-guided processes.

Second, we could attempt the inversion under a null-text condition ($c = c_{\text{null}}$) to bypass this textual defense.
This approach also fails, but in a more subtle way. As shown in Figure~\ref{fig:ddim_inv} (right), the resulting image is visually closer to the original than the text-conditioned attempt, yet it still fails to regenerate the specific target concept (i.e., the artistic style).
This partial failure stems from the standard inversion formula (Eq.~(\ref{eq:ddim_inversion_approx})), which relies on a critical approximation: it uses the noise prediction from the previous step, $\epsilon_{\theta}(z_{t-1}, t-1, c)$, to estimate the next latent $z_t$.
This approximation is heavily dependent on a meaningful text condition $c$ to guide the noise prediction.
In the absence of such guidance ($c_{\text{null}}$), the predictions become unconstrained, and the small error introduced at each step rapidly compounds across the entire trajectory.
Consequently, the final computed noise $\hat{z}_T$ drifts sufficiently from the true required noise $z_T^*$ to lose the high-fidelity details of the target concept, rendering it incapable of an accurate reconstruction.
This demonstrates that standard inversion is an inadequate tool for identifying text-free generative paths.

\begin{algorithm}[t]
\caption{TINA: Text-Free Inversion}
\label{alg:tina_inversion}
\begin{algorithmic}[1]
\State \textbf{Input:} Sanitized denoising model $\epsilon_{\theta}$, target image latent $z_0$, DDIM steps $T$, null-text condition $c_{\text{null}}$, max optimization iterations $K$, learning rate $\eta$.
\State \textbf{Output:} Optimized initial noise $z_T^*$.

\For{$t \leftarrow 1$ to T}
\State Get initial $\tilde{z}_t$ from $z_{t-1}$ using Eq.~(\ref{eq:ddim_inversion_approx}) with $c_{\text{null}}$.
\State $z_t \leftarrow \tilde{z}_t$.

\For{$k \leftarrow 1$ to K}
    \State $\mathcal{L}_t \leftarrow \left\| f_{\theta}^{*}(z_t, z_{t-1}, t, c_{\text{null}}) - z_t \right\|_2^2$ \Comment{Eq.~(\ref{eq:tina_loss})}
    \State $z_t \leftarrow z_t - \eta \nabla_{z_t} \mathcal{L}_t$
\EndFor
\EndFor

\State $z_T^* \leftarrow z_T$
\State \textbf{return} $z_T^*$
\end{algorithmic}
\end{algorithm}

\subsubsection{Optimization-Based Text-Free Inversion}
\label{subsec:solution}

To overcome the imprecision of approximative inversion, TINA reframes the search for the generative trajectory as an optimization problem.
We discard the flawed approximation in Eq.~(\ref{eq:ddim_inversion_approx}) and instead derive the ideal inversion relationship directly from the DDIM sampling equation (Eq.~(\ref{eq:ddim_sample})).
By algebraically rearranging the sampling step, we can express the exact, non-approximative relationship between $z_t$ and $z_{t-1}$:
\begin{equation}
\label{eq:ideal_inversion1}
z_t = C_1(t) z_{t-1} + C_2(t) \cdot \epsilon_\theta(z_{t}, t, c).
\end{equation}
This equation establishes an implicit consistency constraint that must hold at every step of a true generative trajectory. We can therefore cast the exact inversion at step $t$ as a \textbf{fixed-point problem}. In particular, for fixed $z_{t-1}$, $t$, and $c$, any exact latent $z_t$ on the DDIM trajectory must satisfy a self-consistency relation of the form $z_t = f_{\theta}^{*}(z_t, z_{t-1}, t, c)$, i.e., $z_t$ is a fixed point of the induced mapping $f_{\theta}^{*}$ \cite{banach1922operations,ortega2000iterative}. Let us define $f_{\theta}^{*}$ as:
\begin{equation}
\label{eq:fixed_point_func}
f_{\theta}^{*}(z_t, z_{t-1}, t, c) = C_1(t) z_{t-1} + C_2(t) \cdot \epsilon_\theta(z_{t}, t, c).
\end{equation}

Our key innovation is to enforce this self-consistency constraint at each step of the inversion process through optimization, rather than relying on the approximative closed-form update. For each timestep $t$ from $1$ to $T$, given the previously computed $z_{t-1}$, we seek a latent $z_t$ whose induced denoising prediction is consistent with the exact DDIM relation under a null-text condition. We formulate this as the following loss function:
\begin{equation}
\label{eq:tina_loss}
\mathcal{L}_t(z_t) = \left\| f_{\theta}^{*}(z_t, z_{t-1}, t, c_{\mathrm{null}}) - z_t \right\|_2^2,
\end{equation}
where $z_t$ is initialized as $\tilde{z}_t = f_{\theta}(z_{t-1}, t, c_{\mathrm{null}})$ before the optimization.

Instead of accepting the erroneous result from the standard formula, we employ an inner optimization loop at each inversion step $t$. We first obtain an initial estimate for $z_t$ using the conventional DDIM inversion step (Eq.~\ref{eq:ddim_inversion_approx}) under the null-text condition. This estimate then serves as the starting point for an iterative refinement process where we perform gradient descent on $z_t$ to minimize the objective defined in Eq.~(\ref{eq:tina_loss}). In this sense, our optimization can be viewed as seeking a numerically self-consistent point, rather than directly trusting the approximative inversion formula \cite{ortega2000iterative}. This optimization forces each latent $z_t$ along the inversion path to be consistent with the internal dynamics of the sanitized model $\epsilon_{\theta}$ in a text-free setting. The inner loop is executed for a fixed number of iterations. By iteratively applying this optimization-based refinement for $t=1, \dots, T$, we trace a highly accurate path back to the true initial noise $z_T^*$, which is now completely decoupled from any textual influence. The complete procedure for this text-free inversion is formally detailed in Algorithm~\ref{alg:tina_inversion}.

\begin{figure*}[t]
    \centering
    \includegraphics[width=\linewidth]{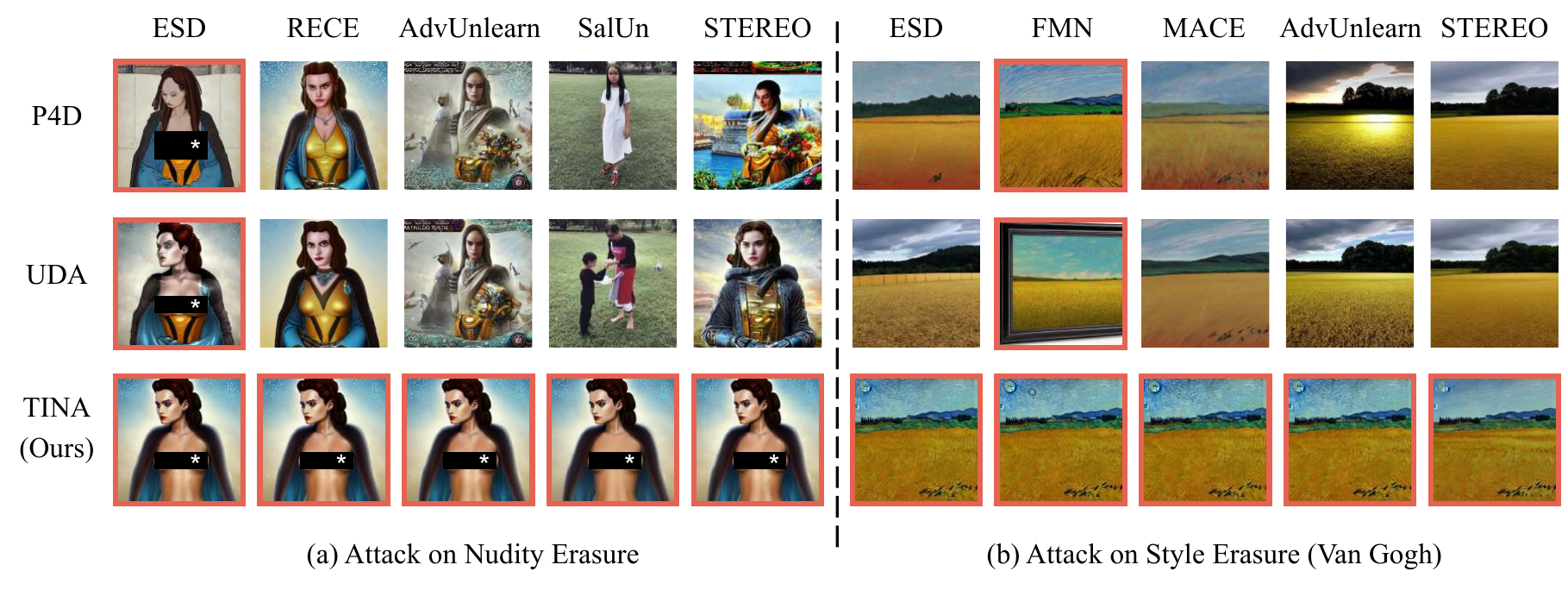}
    \caption{Qualitative comparison of attack performance on (a) Nudity Erasure and (b) Style Erasure (Van Gogh). Images with a \textcolor[RGB]{231, 98, 84}{\textbf{red border}} indicate a successful attack. Our TINA (bottom row) consistently regenerates the forbidden concepts, bypassing most of defenses, while text-centric attacks fail against robust methods. Sensitive content is redacted.}
    \label{fig:visual}
\end{figure*}

\begin{table*}[t]
  \centering
    \scalebox{0.943}{\begin{tabular}{lcccccccc}
    \toprule
          & \textbf{ESD} \cite{gandikota2023erasing} & \textbf{FMN} \cite{zhang2024forget} & \textbf{UCE} \cite{gandikota2024unified} & \textbf{MACE} \cite{lu2024mace} & \textbf{RECE} \cite{gong2024reliable} & \textbf{AdvUnlearn} \cite{zhang2024defensive} & \textbf{SalUn} \cite{fan2023salun} & \textbf{STEREO} \cite{srivatsan2025stereo} \\
    \midrule
    \textbf{MMA} \cite{yang2024mma} & 13.10  & 67.00  & 32.60  & 6.00  & 22.80  & 1.70  & 1.70  & 5.50  \\
    \textbf{P4D} \cite{chin2023prompting4debugging} & 69.01  & 97.89  & 76.06  & 75.35  & 66.20  & 18.31  & 15.49  & 24.65  \\
    \textbf{UDA} \cite{zhang2024generate} & 76.05  & 97.89  & 78.87  & 81.69  & 63.38  & 23.24  & 13.38  & 25.35  \\
    \textbf{RAB} \cite{tsai2023ring} & 50.53  & 97.89  & 29.47  & 6.32  & 10.53  & 2.11  & 0.00  & 8.42  \\
    \textbf{CCE} \cite{pham2023circumventing} & 74.65  & 54.93  & 49.30  & 50.00  & 66.90  & 76.76  & 2.82  & 16.90  \\
    \midrule
    \textbf{TINA} & \textbf{82.39} & \textbf{97.89} & \textbf{82.39} & \textbf{92.96} & \textbf{80.28} & \textbf{78.87} & \textbf{71.13} & \textbf{80.99} \\
    \bottomrule
    \end{tabular}}%
  \caption{Quantitative comparison of Attack Success Rates (ASR, in \%) on the nudity erasure task. We evaluate our TINA against five baselines across eight prominent unlearning defenses. \textbf{Bold} indicates the best-performing attack.}
  \label{tab:nudity}%
\end{table*}%

\section{Experiments}

\subsection{Experimental Setup}

\paragraph{Unlearned DMs to be attacked.}
The field of machine unlearning for diffusion models is advancing at a rapid pace.
For our evaluation, we select target unlearning methods based on two criteria: the public availability of their source code and the reproducibility of their reported results.
Specifically, our evaluation targets twelve prominent, open-source unlearning methods:
(1) ESD (erased stable diffusion) \cite{gandikota2023erasing},
(2) FMN (Forget-Me-Not) \cite{zhang2024forget},
(3) AC (ablating concepts) \cite{kumari2023ablating},
(4) UCE (unified concept editing) \cite{gandikota2024unified},
(5) EraseDiff~\cite{wu2024erasediff},
(6) Scissorhands~\cite{wu2024scissorhands},
(7) MACE (Mass Concept Erasure) \cite{lu2024mace},
(8) SPM (concept-SemiPermeable Membrane) \cite{lyu2024one},
(9) RECE \cite{gong2024reliable},
(10) AdvUnlearn \cite{zhang2024defensive},
(11) SalUn (saliency unlearning) \cite{fan2023salun},
(12) STEREO \cite{srivatsan2025stereo}.
We note that these unlearning methods are often specialized for specific domains (e.g., nudity) rather than being universal.
Consequently, our evaluation is contextualized, assessing each method only on its intended unlearning tasks.

\paragraph{Baseline Attack Methods.}
We evaluate the attack performance of TINA against five prevailing baseline attack methods:
(1) MMA \cite{yang2024mma},
(2) Prompting4Debugging (P4D)~\cite{chin2023prompting4debugging},
(3) UnlearnDiffAtk (UDA) \cite{zhang2024generate},
(4) Ring-A-Bell (RAB) \cite{tsai2023ring},
and (5) CCE \cite{pham2023circumventing}.
The fundamental characteristic and shared limitation of all these baselines is their text-centric nature.
They exclusively target the textual input pathway in an attempt to circumvent textual defenses.
These approaches range from black-box jailbreaking techniques that discover alternative text prompts (e.g., MMA, RAB) to white-box attacks that leverage model access to optimize for adversarial text conditions (e.g., P4D, UDA, CCE).
Critically, all these existing attacks remain tethered to the text-conditioning mechanism, a pathway that TINA is designed to bypass entirely.
As summarized in Table~\ref{tab:baseline_applicability}, not all baseline attacks are designed for every type of concept (e.g., MMA and RAB are not applicable to style or object erasure).
Therefore, our comparative analysis is conducted on a per-task basis, aligning with the intended application of each attack method.
The implementing details of these baselines can be found in Appendix.~\ref{sec:attack_setup}.

\begin{table*}[t] 
  \centering
    \scalebox{0.95}{\begin{tabular}{lcccccccc}
    \toprule
          & \textbf{ESD} \cite{gandikota2023erasing}   & \textbf{FMN} \cite{zhang2024forget}   & \textbf{AC} \cite{kumari2023ablating}   & \textbf{MACE} \cite{lu2024mace} & \textbf{SPM} \cite{lyu2024one} & \textbf{RECE} \cite{gong2024reliable} & \textbf{AdvUnlearn} \cite{zhang2024defensive} & \textbf{STEREO} \cite{srivatsan2025stereo} \\
    \midrule
    \textbf{P4D} \cite{chin2023prompting4debugging}   & 30.0  & 54.0  & 68.0  & 42.0  & 78.0  & 62.0  & 0.0   & 0.0  \\
    \textbf{UDA} \cite{zhang2024generate} & \underline{32.0}  & \underline{56.0}  & \textbf{77.0} & \underline{56.0}  & \textbf{88.0} & \underline{64.0}  & 2.0   & 0.0  \\
    \textbf{CCE} \cite{pham2023circumventing}   & 8.0   & 18.0  & 14.0  & 26.0  & 36.0  & 40.0  & \underline{44.0}  & \underline{4.0}  \\
    \midrule
    \textbf{TINA}  & \textbf{70.0} & \textbf{72.0} & \underline{74.0}  & \textbf{72.0} & \underline{80.0}  & \textbf{74.0} & \textbf{70.0} & \textbf{44.0} \\
    \bottomrule
    \end{tabular}}%
  \caption{Comparison of Attack Success Rates (ASR, in \%) on the artistic style erasure task. We report the performance of our TINA against three baselines across eight unlearning methods. \textbf{Bold} denotes the highest ASR, while \underline{underlined} denotes the second highest.}
  \label{tab:style}
\end{table*}

\begin{table}[t]
    \centering
    \begin{tabular}{ccccccc}
    \toprule
            & \textbf{MMA} & \textbf{P4D} & \textbf{UDA} & \textbf{RAB} & \textbf{CCE} \\
    \midrule
    \textbf{Nudity} & \textcolor{green}{\checkmark}     & \textcolor{green}{\checkmark}     & \textcolor{green}{\checkmark}     & \textcolor{green}{\checkmark}     & \textcolor{green}{\checkmark}     \\
    \textbf{Style} & \textcolor{red}{\ding{55}}     & \textcolor{green}{\checkmark}     & \textcolor{green}{\checkmark}     & \textcolor{red}{\ding{55}}     & \textcolor{green}{\checkmark}     \\
    \textbf{Object} & \textcolor{red}{\ding{55}}     & \textcolor{green}{\checkmark}     & \textcolor{green}{\checkmark}     & \textcolor{red}{\ding{55}}     & \textcolor{green}{\checkmark}     \\
    \bottomrule
    \end{tabular}%
    \caption{Applicability of baseline attack methods to different tasks. \textcolor{green}{\checkmark} denotes applicability, while \textcolor{red}{\ding{55}} denotes non-applicability.}
    \label{tab:baseline_applicability}%
\end{table}%

\paragraph{Image Setup.}
Our method requires a set of representative images, each exemplifying a concept targeted for erasure.
We obtain these images by adopting the generation setup established in the UnlearnDiffAtk (UDA) study~\cite{zhang2024generate}.
This process involves using the original Stable Diffusion v1.4 model, which is the pre-trained checkpoint before any unlearning methods were applied, to generate a collection of ground-truth images.
The generation is guided by text prompts sourced from the same standard benchmarks used to evaluate the baseline attacks.
Specifically, for nudity concepts, we use prompts from the I2P dataset~\cite{schramowski2023safe}.
For artistic styles, we focus our experiments on the Van Gogh style, using art-related prompts from the ESD evaluation setup~\cite{gandikota2023erasing}.
For objects, we concentrate on the tench class, using prompts generated via GPT-4 corresponding to this Imagenette class, following the methodology in~\cite{zhang2024generate}.

\paragraph{Implementation Details.}
All experiments are built upon the Stable Diffusion v1.4 checkpoint.
For all generative processes, we employ a linear multistep scheduler with $T=50$ inference steps and a standard Classifier-Free Guidance (CFG~\cite{ho2022classifier}) scale of 7.5.
For our proposed TINA, the optimization-based inversion (detailed in Section~\ref{sec:tina} and Algorithm~\ref{alg:tina_inversion}) is also configured with $T=50$ total steps. Within the inner optimization loop at each timestep $t$, we perform $K=25$ refinement iterations to minimize the fixed-point loss (Eq.~\ref{eq:tina_loss}).
This inner optimization uses the AdamW~\cite{loshchilov2017decoupled} with a learning rate of $\eta = 0.001$.
All experiments were conducted on a single NVIDIA A100 GPU.

\paragraph{Evaluation Metrics.}
Our evaluation methodology is primarily aligned with the protocol established by UnlearnDiffAtk (UDA)~\cite{zhang2024generate}.
To quantitatively measure attack efficacy, we report the Attack Success Rate (\textbf{ASR}), defined as the percentage of generated images that are successfully identified by a post-generation classifier as containing the forbidden concept, thereby indicating a successful bypass of the unlearning safeguard.
We employ a suite of specialized classifiers corresponding to each unlearning task.
For harmful concept unlearning, we utilize NudeNet to detect nudity.
For style unlearning, we adopt the 129-class ViT-base style classifier~\cite{wu2020visual,zhang2024generate}, which was pre-trained on ImageNet, fine-tuned on the WikiArt dataset~\cite{saleh2015large}.
For object unlearning, we employ a standard ImageNet-pretrained ResNet-50~\cite{he2016deep} for generated image classification.

\subsection{Experiment Results}
\label{sec:results}

We now empirically validate the efficacy of TINA.
Our experiments are designed to test the ability of TINA to bypass a wide spectrum of unlearning defenses, from foundational methods to the current state-of-the-art, and compare its performance directly against existing text-centric attacks.
The results for nudity, style, and object erasure are summarized in Tables \ref{tab:nudity}, \ref{tab:style}, and \ref{tab:object}, respectively.
Qualitative comparisons and an ablation study are provided in Figure~\ref{fig:visual}, Section~\ref{sec:more_visual_results}, and Appendix~\ref{sec:appx_ablation}.

\paragraph{Evaluation on Erased Models in Nudity Erasure.}
Table~\ref{tab:nudity} demonstrates comprehensive success of TINA in the nudity erasure task.
TINA achieves the highest ASR across all eight defenses.
It reaches $82.39\%$ against ESD and $92.96\%$ against MACE, significantly outperforming all baselines.
The most critical finding is the performance of TINA against defenses designed to be robust, such as AdvUnlearn, SalUn, and STEREO.
Against these targets, text-based attacks are largely mitigated (e.g., UDA scores $23.24\%$ on AdvUnlearn, RAB scores $0.00\%$ on SalUn).
In contrast, TINA maintains high ASRs ($78.87\%$, $71.13\%$, and $80.99\%$), proving it exploits a vulnerability that these defenses do not account for.
This quantitative success is visually corroborated in Figure~\ref{fig:visual}~(a).
The bottom row shows TINA consistently regenerating the target content across all defenses.
In contrast, text-centric attacks like P4D and UDA visibly fail against robust methods such as AdvUnlearn and STEREO, producing neutral or unrelated images.

\paragraph{Evaluation on Erased Models in Style Erasure.}
The results for artistic style erasure (Table~\ref{tab:style}) further reinforce this trend.
TINA robustly bypasses the majority of defenses. Its $70.0\%$ ASR against ESD starkly contrasts with the $32.0\%$ from UDA and $8.0\%$ from CCE.
Furthermore, TINA shows dominant performance against robust methods like AdvUnlearn ($70.0\%$) and STEREO ($44.0\%$), where text-based attacks remain weak or ineffective.
This finding demonstrates that even for abstract concepts like style, a deterministic, non-textual generative trajectory persists after erasure, which TINA successfully identifies and exploits.
Figure~\ref{fig:visual}~(b) provides the qualitative evidence for this trend. It clearly demonstrates the ability of TINA to restore the ``Van Gogh'' style, bypassing defenses like MACE and AdvUnlearn where text-based attacks fail to regenerate the artistic features.

\paragraph{Evaluation on Erased Models in Object Erasure.}
The object erasure evaluation in Table~\ref{tab:object} delivers the most compelling results.
This task highlights the failure of text-centric unlearning against our text-free attack.
Against modern defenses like Scissorhands and STEREO, all text-based attacks (P4D, UDA, and CCE) fail catastrophically, with ASRs in the single digits.
While P4D and UDA also fail against EraseDiff and SalUn, the embedding-space attack CCE retains partial efficacy (34.0\% and 58.0\%, respectively).
This pattern confirms that while robust defenses are patching textual vulnerabilities, TINA remains unaffected.
It achieves uniformly high ASRs across all targets, including 68.0\% (EraseDiff), 78.0\% (Scissorhands), and 72.0\% (STEREO).
This finding confirms our central hypothesis: current SOTA methods merely sever text-image links, not the underlying visual knowledge, which TINA can deterministically access without any textual guidance.

\begin{table}[t]
  \centering
    \begin{tabular}{lcccc}
    \toprule
          & \textbf{P4D} & \textbf{UDA} & \textbf{CCE} & \textbf{TINA} \\
    \midrule
    \textbf{ESD} \cite{gandikota2023erasing} & 32.0  & 46.0 & 40.0 & \textbf{70.0} \\
    \textbf{EraseDiff} \cite{wu2024erasediff} & 8.0   & 2.0  &  34.0  & \textbf{68.0} \\
    \textbf{SalUn} \cite{fan2023salun} & 18.0  & 12.0 & 58.0 & \textbf{72.0} \\
    \textbf{Scissorhands} \cite{wu2024scissorhands} & 6.0   & 6.0 & 0.0  & \textbf{78.0} \\
    \textbf{STEREO} \cite{srivatsan2025stereo} & 0.0   & 2.0  & 2.0 & \textbf{72.0} \\
    \bottomrule
    \end{tabular}%
  \caption{Attack Success Rates (ASR, in \%) for object erasure. Our TINA bypasses modern defenses where text-centric attacks fail.}
  \label{tab:object}%
\end{table}%

\subsection{Latent Embedding Analysis}
\begin{figure}[t]
    \centering
    \includegraphics[width=\linewidth]{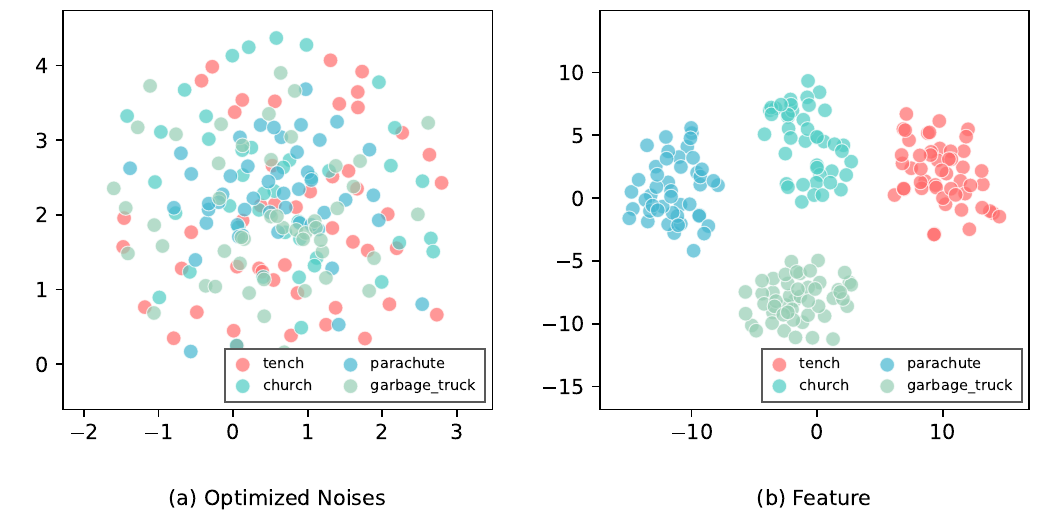}
    \caption{t-SNE visualization of (a) the optimized initial noises $z_T^*$ and (b) their corresponding deep UNet activations (extracted from the \texttt{mid\_block}) for four erased concepts.}
    \label{fig:tsne}
\end{figure}

To further understand the generative pathway identified by TINA, we visualize the learned noise embeddings $z_T^*$ for four erased concepts (Tench, Church, Parachute, Garbage Truck) using a t-SNE projection.
Specifically, for each concept, we collect 50 optimized noise vectors generated by targeting the corresponding ESD-sanitized model.
We then feed these optimized noises into the UNet under a null-text condition and extract their intermediate representations through a forward inference. 
For the visualization, we apply t-SNE (perplexity 30, with PCA initialization) to both the flattened noise vectors and the global average pooled features extracted from the last ResNet block of the \texttt{mid\_block} in the UNet model.
As shown in Figure~\ref{fig:tsne}, the optimized noises themselves (left) exhibit little concept-wise separation in the latent space, which is consistent with their noise-like nature.
However, their corresponding internal UNet activations (right) become clearly separable by concept, forming distinct clusters.
This demonstrates that although the optimized $z_T^*$ remains indistinguishable from random noise, it successfully elicits concept-specific internal responses within the erased diffusion model. This confirms that the forbidden visual knowledge persists and is deterministically activated by our text-free attack.

\section{Conclusion}
We exposed a fundamental flaw in the text-centric paradigm of concept erasure: severing text-image links is insufficient and offers a false sense of security.
We introduced TINA, a novel text-free attack that substantiates this claim by completely bypassing textual controls.
TINA identifies a deterministic, null-text generative trajectory to regenerate forbidden content.
Our experiments show that TINA consistently restores erased concepts from models sanitized by existing defenses.
This proves current methods merely obscure visual knowledge rather than erasing it, mandating a shift towards unlearning techniques that operate directly on internal visual representations.

\clearpage

{
    \small
    \bibliographystyle{ieeenat_fullname}
    \bibliography{main}

@inproceedings{rombach2022high,
  title={High-resolution image synthesis with latent diffusion models},
  author={Rombach, Robin and Blattmann, Andreas and Lorenz, Dominik and Esser, Patrick and Ommer, Bj{\"o}rn},
  booktitle={Proceedings of the IEEE/CVF conference on computer vision and pattern recognition},
  pages={10684--10695},
  year={2022}
}

@article{schuhmann2022laion,
  title={Laion-5b: An open large-scale dataset for training next generation image-text models},
  author={Schuhmann, Christoph and Beaumont, Romain and Vencu, Richard and Gordon, Cade and Wightman, Ross and Cherti, Mehdi and Coombes, Theo and Katta, Aarush and Mullis, Clayton and Wortsman, Mitchell and others},
  journal={Advances in neural information processing systems},
  volume={35},
  pages={25278--25294},
  year={2022}
}

@inproceedings{jiang2023ai,
  title={AI Art and its Impact on Artists},
  author={Jiang, Harry H and Brown, Lauren and Cheng, Jessica and Khan, Mehtab and Gupta, Abhishek and Workman, Deja and Hanna, Alex and Flowers, Johnathan and Gebru, Timnit},
  booktitle={Proceedings of the 2023 AAAI/ACM Conference on AI, Ethics, and Society},
  pages={363--374},
  year={2023}
}

@inproceedings{zhang2024generate,
  title={To generate or not? safety-driven unlearned diffusion models are still easy to generate unsafe images... for now},
  author={Zhang, Yimeng and Jia, Jinghan and Chen, Xin and Chen, Aochuan and Zhang, Yihua and Liu, Jiancheng and Ding, Ke and Liu, Sijia},
  booktitle={European Conference on Computer Vision},
  pages={385--403},
  year={2024},
  organization={Springer}
}

@article{nichol2021glide,
  title={Glide: Towards photorealistic image generation and editing with text-guided diffusion models},
  author={Nichol, Alex and Dhariwal, Prafulla and Ramesh, Aditya and Shyam, Pranav and Mishkin, Pamela and McGrew, Bob and Sutskever, Ilya and Chen, Mark},
  journal={arXiv preprint arXiv:2112.10741},
  year={2021}
}

@inproceedings{ramesh2021zero,
  title={Zero-shot text-to-image generation},
  author={Ramesh, Aditya and Pavlov, Mikhail and Goh, Gabriel and Gray, Scott and Voss, Chelsea and Radford, Alec and Chen, Mark and Sutskever, Ilya},
  booktitle={International conference on machine learning},
  pages={8821--8831},
  year={2021},
  organization={Pmlr}
}

@article{saharia2022photorealistic,
  title={Photorealistic text-to-image diffusion models with deep language understanding},
  author={Saharia, Chitwan and Chan, William and Saxena, Saurabh and Li, Lala and Whang, Jay and Denton, Emily L and Ghasemipour, Kamyar and Gontijo Lopes, Raphael and Karagol Ayan, Burcu and Salimans, Tim and others},
  journal={Advances in neural information processing systems},
  volume={35},
  pages={36479--36494},
  year={2022}
}

@article{roose2022ai,
  title={An AI-Generated Picture Won an Art Prize. Artists Aren't Happy.},
  author={Roose, Kevin},
  journal={New York Times},
  volume={16},
  number={01},
  pages={2025},
  year={2022}
}

@article{hunter2023ai,
  title={AI porn is easy to make now. For women, that's a nightmare.},
  author={Hunter, Tatum},
  journal={The Washington Post},
  pages={NA--NA},
  year={2023},
  publisher={The Washington Post}
}

@inproceedings{carlini2023extracting,
  title={Extracting training data from diffusion models},
  author={Carlini, Nicolas and Hayes, Jamie and Nasr, Milad and Jagielski, Matthew and Sehwag, Vikash and Tramer, Florian and Balle, Borja and Ippolito, Daphne and Wallace, Eric},
  booktitle={32nd USENIX security symposium (USENIX Security 23)},
  pages={5253--5270},
  year={2023}
}

@inproceedings{somepalli2023diffusion,
  title={Diffusion art or digital forgery? investigating data replication in diffusion models},
  author={Somepalli, Gowthami and Singla, Vasu and Goldblum, Micah and Geiping, Jonas and Goldstein, Tom},
  booktitle={Proceedings of the IEEE/CVF conference on computer vision and pattern recognition},
  pages={6048--6058},
  year={2023}
}

@inproceedings{schramowski2023safe,
  title={Safe latent diffusion: Mitigating inappropriate degeneration in diffusion models},
  author={Schramowski, Patrick and Brack, Manuel and Deiseroth, Bj{\"o}rn and Kersting, Kristian},
  booktitle={Proceedings of the IEEE/CVF Conference on Computer Vision and Pattern Recognition},
  pages={22522--22531},
  year={2023}
}

@inproceedings{gandikota2023erasing,
  title={Erasing concepts from diffusion models},
  author={Gandikota, Rohit and Materzynska, Joanna and Fiotto-Kaufman, Jaden and Bau, David},
  booktitle={Proceedings of the IEEE/CVF international conference on computer vision},
  pages={2426--2436},
  year={2023}
}

@inproceedings{kumari2023ablating,
  title={Ablating concepts in text-to-image diffusion models},
  author={Kumari, Nupur and Zhang, Bingliang and Wang, Sheng-Yu and Shechtman, Eli and Zhang, Richard and Zhu, Jun-Yan},
  booktitle={Proceedings of the IEEE/CVF International Conference on Computer Vision},
  pages={22691--22702},
  year={2023}
}

@article{xu2023machine,
    title = {Machine Unlearning: A Survey},
    author = {Xu, Heng and Zhu, Tianqing and Zhang, Lefeng and Zhou, Wanlei and Yu, Philip S.},
    year = {2023},
    issue_date = {January 2024},
    publisher = {Association for Computing Machinery},
    address = {New York, NY, USA},
    volume = {56},
    number = {1},
    issn = {0360-0300},
    url = {https://doi.org/10.1145/3603620},
    doi = {10.1145/3603620},
    journal = {ACM Comput. Surv.},
    month = aug,
    articleno = {9},
    numpages = {36},
}

@inproceedings{zhang2024forget,
  title={Forget-me-not: Learning to forget in text-to-image diffusion models},
  author={Zhang, Gong and Wang, Kai and Xu, Xingqian and Wang, Zhangyang and Shi, Humphrey},
  booktitle={Proceedings of the IEEE/CVF conference on computer vision and pattern recognition},
  pages={1755--1764},
  year={2024}
}

@inproceedings{gandikota2024unified,
  title={Unified concept editing in diffusion models},
  author={Gandikota, Rohit and Orgad, Hadas and Belinkov, Yonatan and Materzy{\'n}ska, Joanna and Bau, David},
  booktitle={Proceedings of the IEEE/CVF Winter Conference on Applications of Computer Vision},
  pages={5111--5120},
  year={2024}
}

@article{chin2023prompting4debugging,
  title={Prompting4debugging: Red-teaming text-to-image diffusion models by finding problematic prompts},
  author={Chin, Zhi-Yi and Jiang, Chieh-Ming and Huang, Ching-Chun and Chen, Pin-Yu and Chiu, Wei-Chen},
  journal={arXiv preprint arXiv:2309.06135},
  year={2023}
}

@article{tsai2023ring,
  title={Ring-a-bell! how reliable are concept removal methods for diffusion models?},
  author={Tsai, Yu-Lin and Hsu, Chia-Yi and Xie, Chulin and Lin, Chih-Hsun and Chen, Jia-You and Li, Bo and Chen, Pin-Yu and Yu, Chia-Mu and Huang, Chun-Ying},
  journal={arXiv preprint arXiv:2310.10012},
  year={2023}
}

@article{pham2023circumventing,
  title={Circumventing concept erasure methods for text-to-image generative models},
  author={Pham, Minh and Marshall, Kelly O and Cohen, Niv and Mittal, Govind and Hegde, Chinmay},
  journal={arXiv preprint arXiv:2308.01508},
  year={2023}
}

@article{song2020denoising,
  title={Denoising diffusion implicit models},
  author={Song, Jiaming and Meng, Chenlin and Ermon, Stefano},
  journal={arXiv preprint arXiv:2010.02502},
  year={2020}
}

@article{ramesh2022hierarchical,
  title={Hierarchical text-conditional image generation with clip latents},
  author={Ramesh, Aditya and Dhariwal, Prafulla and Nichol, Alex and Chu, Casey and Chen, Mark},
  journal={arXiv preprint arXiv:2204.06125},
  volume={1},
  number={2},
  pages={3},
  year={2022}
}

@article{ho2020denoising,
  title={Denoising diffusion probabilistic models},
  author={Ho, Jonathan and Jain, Ajay and Abbeel, Pieter},
  journal={Advances in neural information processing systems},
  volume={33},
  pages={6840--6851},
  year={2020}
}

@article{vaswani2017attention,
  title={Attention is all you need},
  author={Vaswani, Ashish and Shazeer, Noam and Parmar, Niki and Uszkoreit, Jakob and Jones, Llion and Gomez, Aidan N and Kaiser, {\L}ukasz and Polosukhin, Illia},
  journal={Advances in neural information processing systems},
  volume={30},
  year={2017}
}

@inproceedings{lu2024mace,
  title={Mace: Mass concept erasure in diffusion models},
  author={Lu, Shilin and Wang, Zilan and Li, Leyang and Liu, Yanzhu and Kong, Adams Wai-Kin},
  booktitle={Proceedings of the IEEE/CVF Conference on Computer Vision and Pattern Recognition},
  pages={6430--6440},
  year={2024}
}

@inproceedings{lyu2024one,
  title={One-dimensional adapter to rule them all: Concepts diffusion models and erasing applications},
  author={Lyu, Mengyao and Yang, Yuhong and Hong, Haiwen and Chen, Hui and Jin, Xuan and He, Yuan and Xue, Hui and Han, Jungong and Ding, Guiguang},
  booktitle={Proceedings of the IEEE/CVF Conference on Computer Vision and Pattern Recognition},
  pages={7559--7568},
  year={2024}
}

@inproceedings{gong2024reliable,
  title={Reliable and efficient concept erasure of text-to-image diffusion models},
  author={Gong, Chao and Chen, Kai and Wei, Zhipeng and Chen, Jingjing and Jiang, Yu-Gang},
  booktitle={European Conference on Computer Vision},
  pages={73--88},
  year={2024},
  organization={Springer}
}

@article{zhang2024defensive,
  title={Defensive unlearning with adversarial training for robust concept erasure in diffusion models},
  author={Zhang, Yimeng and Chen, Xin and Jia, Jinghan and Zhang, Yihua and Fan, Chongyu and Liu, Jiancheng and Hong, Mingyi and Ding, Ke and Liu, Sijia},
  journal={Advances in neural information processing systems},
  volume={37},
  pages={36748--36776},
  year={2024}
}

@article{fan2023salun,
  title={Salun: Empowering machine unlearning via gradient-based weight saliency in both image classification and generation},
  author={Fan, Chongyu and Liu, Jiancheng and Zhang, Yihua and Wong, Eric and Wei, Dennis and Liu, Sijia},
  journal={arXiv preprint arXiv:2310.12508},
  year={2023}
}

@inproceedings{srivatsan2025stereo,
  title={Stereo: A two-stage framework for adversarially robust concept erasing from text-to-image diffusion models},
  author={Srivatsan, Koushik and Shamshad, Fahad and Naseer, Muzammal and Patel, Vishal M and Nandakumar, Karthik},
  booktitle={Proceedings of the Computer Vision and Pattern Recognition Conference},
  pages={23765--23774},
  year={2025}
}

@inproceedings{yang2024mma,
  title={Mma-diffusion: Multimodal attack on diffusion models},
  author={Yang, Yijun and Gao, Ruiyuan and Wang, Xiaosen and Ho, Tsung-Yi and Xu, Nan and Xu, Qiang},
  booktitle={Proceedings of the IEEE/CVF Conference on Computer Vision and Pattern Recognition},
  pages={7737--7746},
  year={2024}
}

@article{wu2024erasediff,
  title={Erasediff: Erasing data influence in diffusion models},
  author={Wu, Jing and Le, Trung and Hayat, Munawar and Harandi, Mehrtash},
  journal={arXiv preprint arXiv:2401.05779},
  year={2024}
}

@inproceedings{wu2024scissorhands,
  title={Scissorhands: Scrub data influence via connection sensitivity in networks},
  author={Wu, Jing and Harandi, Mehrtash},
  booktitle={European Conference on Computer Vision},
  pages={367--384},
  year={2024},
  organization={Springer}
}

@article{loshchilov2017decoupled,
  title={Decoupled weight decay regularization},
  author={Loshchilov, Ilya and Hutter, Frank},
  journal={arXiv preprint arXiv:1711.05101},
  year={2017}
}

@article{ho2022classifier,
  title={Classifier-free diffusion guidance},
  author={Ho, Jonathan and Salimans, Tim},
  journal={arXiv preprint arXiv:2207.12598},
  year={2022}
}

@article{wu2020visual,
  title={Visual transformers: Token-based image representation and processing for computer vision},
  author={Wu, Bichen and Xu, Chenfeng and Dai, Xiaoliang and Wan, Alvin and Zhang, Peizhao and Yan, Zhicheng and Tomizuka, Masayoshi and Gonzalez, Joseph and Keutzer, Kurt and Vajda, Peter},
  journal={arXiv preprint arXiv:2006.03677},
  year={2020}
}

@article{saleh2015large,
  title={Large-scale classification of fine-art paintings: Learning the right metric on the right feature},
  author={Saleh, Babak and Elgammal, Ahmed},
  journal={arXiv preprint arXiv:1505.00855},
  year={2015}
}

@inproceedings{he2016deep,
  title={Deep residual learning for image recognition},
  author={He, Kaiming and Zhang, Xiangyu and Ren, Shaoqing and Sun, Jian},
  booktitle={Proceedings of the IEEE conference on computer vision and pattern recognition},
  pages={770--778},
  year={2016}
}

@inproceedings{peebles2023scalable,
  title={Scalable diffusion models with transformers},
  author={Peebles, William and Xie, Saining},
  booktitle={Proceedings of the IEEE/CVF international conference on computer vision},
  pages={4195--4205},
  year={2023}
}

@inproceedings{chen2023pixart,
  title={PixArt-\${\textbackslash}alpha\$: Fast Training of Diffusion Transformer for Photorealistic Text-to-Image Synthesis},
  author={Junsong Chen and Jincheng YU and Chongjian GE and Lewei Yao and Enze Xie and Zhongdao Wang and James Kwok and Ping Luo and Huchuan Lu and Zhenguo Li},
  booktitle={The Twelfth International Conference on Learning Representations},
  year={2024},
  url={https://openreview.net/forum?id=eAKmQPe3m1}
}

@inproceedings{zhang2025easyinv,
  title={EasyInv: Toward Fast and Better DDIM Inversion},
  author={Ziyue Zhang and Mingbao Lin and Shuicheng YAN and Rongrong Ji},
  booktitle={Forty-second International Conference on Machine Learning},
  year={2025},
}

@inproceedings{mokady2023null,
  title={Null-text inversion for editing real images using guided diffusion models},
  author={Mokady, Ron and Hertz, Amir and Aberman, Kfir and Pritch, Yael and Cohen-Or, Daniel},
  booktitle={Proceedings of the IEEE/CVF conference on computer vision and pattern recognition},
  pages={6038--6047},
  year={2023}
}

@inproceedings{miyake2025negative,
  title={Negative-prompt inversion: Fast image inversion for editing with text-guided diffusion models},
  author={Miyake, Daiki and Iohara, Akihiro and Saito, Yu and Tanaka, Toshiyuki},
  booktitle={2025 IEEE/CVF Winter Conference on Applications of Computer Vision (WACV)},
  pages={2063--2072},
  year={2025},
  organization={IEEE}
}

@inproceedings{ju2023pnp,
  title={Pnp inversion: Boosting diffusion-based editing with 3 lines of code},
  author={Ju, Xuan and Zeng, Ailing and Bian, Yuxuan and Liu, Shaoteng and Xu, Qiang},
  booktitle={The Twelfth International Conference on Learning Representations},
  year={2023}
}

@article{banach1922operations,
  title={Sur les op{\'e}rations dans les ensembles abstraits et leur application aux {\'e}quations int{\'e}grales},
  author={Banach, Stefan},
  journal={Fundamenta mathematicae},
  volume={3},
  number={1},
  pages={133--181},
  year={1922},
  publisher={Polska Akademia Nauk. Instytut Matematyczny PAN}
}

@book{ortega2000iterative,
  title={Iterative solution of nonlinear equations in several variables},
  author={Ortega, James M and Rheinboldt, Werner C},
  year={2000},
  publisher={SIAM}
}

@inproceedings{zhang2024multi,
  title={Multi-factor adaptive vision selection for egocentric video question answering},
  author={Zhang, Haoyu and Liu, Meng and Liu, Zixin and Song, Xuemeng and Wang, Yaowei and Nie, Liqiang},
  booktitle={Forty-first International Conference on Machine Learning},
  pages={59310--59328},
  year={2024},
  volume={235}
}

@inproceedings{zhang2025spatial,
  title={Spatial Understanding from Videos: Structured Prompts Meet Simulation Data},
  author={Zhang, Haoyu and Liu, Meng and Li, Zaijing and Wen, Haokun and Guan, Weili and Wang, Yaowei and Nie, Liqiang},
  booktitle={Advances in Neural Information Processing Systems},
  year={2025},
  pages={1-16}
}

@article{zhang2023attribute,
  title={Attribute-guided collaborative learning for partial person re-identification},
  author={Zhang, Haoyu and Liu, Meng and Li, Yuhong and Yan, Ming and Gao, Zan and Chang, Xiaojun and Nie, Liqiang},
  journal={IEEE Transactions on Pattern Analysis and Machine Intelligence},
  volume={45},
  number={12},
  pages={14144--14160},
  year={2023},
  publisher={IEEE}
}

@article{li2026optimus3,
  title={Optimus-3: Dual-Router Aligned Mixture-of-Experts Agent with Dual-Granularity Reasoning-Aware Policy Optimization},
  author={Li, Zaijing and Xie, Yuquan and Shao, Rui and Chen, Gongwei and Guan, Weili and Jiang, Dongmei and Wang, Yaowei and Nie, Liqiang},
  journal={arXiv preprint arXiv:2506.10357},
  year={2025}
}

@article{li2026optimusvla,
  title={Global Prior Meets Local Consistency: Dual-Memory Augmented Vision-Language-Action Model for Efficient Robotic Manipulation},
  author={Li, Zaijing and Hu, Bing and Shao, Rui and Chen, Gongwei and Jiang, Dongmei and Xie, Pengwei and Hao, Jianye and Nie, Liqiang},
  journal={arXiv preprint arXiv:2602.20200},
  year={2026}
}

@inproceedings{wang2025unifying,
  title={Unifying Reconstruction and Density Estimation via Invertible Contraction Mapping in One-Class Classification},
  author={Wang, Xiaolei and Dai, Tianhong and Bai, Huihui and Zhao, Yao and Xiao, Jimin},
  booktitle={The Thirty-ninth Annual Conference on Neural Information Processing Systems},
  year={2025},
}

@inproceedings{wang2025decad,
  title={Decad: Decoupling anomalies in latent space for multi-class unsupervised anomaly detection},
  author={Wang, Xiaolei and Wang, Xiaoyang and Bai, Huihui and Lim, Eng Gee and Xiao, Jimin},
  booktitle={Proceedings of the IEEE/CVF International Conference on Computer Vision},
  pages={21568--21577},
  year={2025}
}

@article{liu2025understanding,
  title={Understanding before recommendation: Semantic aspect-aware review exploitation via large language models},
  author={Liu, Fan and Liu, Yaqi and Chen, Huilin and Cheng, Zhiyong and Nie, Liqiang and Kankanhalli, Mohan},
  journal={ACM Transactions on Information Systems},
  volume={43},
  number={2},
  pages={1--26},
  year={2025},
  publisher={ACM New York, NY}
}

@article{wang2026cross,
  title={Cross-Modal Representation Shift Refinement for Point-supervised Video Moment Retrieval},
  author={Wang, Kun and Hu, Yupeng and Liu, Hao and Shao, Jiang and Nie, Liqiang},
  journal={ACM Transactions on Information Systems},
  volume={44},
  number={3},
  pages={1--30},
  year={2026},
  publisher={ACM New York, NY}
}

@inproceedings{wang2024explicit,
  title={Explicit Granularity and Implicit Scale Correspondence Learning for Point-Supervised Video Moment Localization},
  author={Wang, Kun and Liu, Hao and Jie, Lirong and Li, Zixu and Hu, Yupeng and Nie, Liqiang},
  booktitle={Proceedings of the 32nd ACM International Conference on Multimedia},
  pages={9214--9223},
  year={2024}
}

@article{wang2025redundancy,
  title={Redundancy Mitigation: Towards Accurate and Efficient Image-Text Retrieval},
  author={Wang, Kun and Hu, Yupeng and Liu, Hao and Jie, Lirong and Nie, Liqiang},
  journal={IEEE Transactions on Circuits and Systems for Video Technology},
  year={2025},
  publisher={IEEE}
}

@inproceedings{xiang2025dkdm,
  title={Dkdm: Data-free knowledge distillation for diffusion models with any architecture},
  author={Xiang, Qianlong and Zhang, Miao and Shang, Yuzhang and Wu, Jianlong and Yan, Yan and Nie, Liqiang},
  booktitle={Proceedings of the IEEE/CVF Conference on Computer Vision and Pattern Recognition},
  pages={2955--2965},
  year={2025}
}
}


\clearpage
\maketitlesupplementary
\section{Baseline Attack Setup}
\label{sec:attack_setup}

We evaluate the robustness of the proposed method against five existing attack methods: MMA~\cite{yang2024mma}, Prompting4Debugging (P4D)~\cite{chin2023prompting4debugging}, UnlearnDiffAtk (UDA)~\cite{zhang2024generate}, Ring-A-Bell (RAB)~\cite{tsai2023ring} and CCE~\cite{pham2023circumventing}. The details are presented below.

\begin{itemize}
\item \textbf{MMA}~\cite{yang2024mma}: This method provides a publicly available set of 1,000 adversarial prompts, curated to bypass safeguards and generate nudity content, specifically targeting Stable Diffusion v1.5.
Given that these prompts are designed for NSFW generation, we employ this dataset as a black-box attack to evaluate robustness on the nudity erasure task.
For each unlearned diffusion model, our protocol involves using these 1,000 prompts as text conditions to generate 1,000 corresponding images.
A fixed random seed is used for each generation to ensure reproducibility. The resulting images are subsequently evaluated using NudeNet to detect nudity and calculate the Attack Success Rate (ASR).
\item \textbf{UDA} (UnlearnDiffAtk)~\cite{zhang2024generate}: This white-box attack aims to find a prompt that minimizes the denoising error of the sanitized model with respect to images containing the forbidden concepts.
To ensure a fair comparison, we strictly follow the original attack setup. For the nudity task, we use 142 prompts from the I2P dataset~\cite{schramowski2023safe} and set the number of optimized tokens to $N=5$.
For the artistic style (Van Gogh) and object (tench) tasks, we use 50 prompts each, with $N=3$ optimized tokens. In all tasks, the adversarial perturbations are optimized for 40 iterations over 50 diffusion timesteps, using the AdamW optimizer with a 0.01 learning rate, consistent with the methodology in~\cite{zhang2024generate}.
\item \textbf{P4D} (Prompting4Debugging)~\cite{chin2023prompting4debugging}: This white-box attack requires access to both the original vanilla model and the target sanitized model.
Its objective is to optimize an adversarial prompt $c'$ that forces the sanitized model's noise prediction to match the original prediction of the model under the forbidden prompt $c$.
To ensure a fair comparison, our P4D implementation strictly follows the experimental setup of UDA, using the identical prompt sets, the same number of optimized tokens ($N=5$ for nudity, $N=3$ for style/object), and the same optimization parameters (40 iterations, AdamW optimizer, 0.01 learning rate).
\item \textbf{RAB} (Ring-A-Bell)~\cite{tsai2023ring}: This attack method is specifically designed to generate adversarial prompts for bypassing nudity safeguards.
Therefore, we evaluate this attack exclusively on the nudity erasure task. We employ the 95 adversarial prompts released by the official RAB implementation.
As detailed in the original paper~\cite{tsai2023ring}, this set of 95 prompts is generated by modifying the base I2P nudity prompts using the recommended hyperparameters for this task: a length of prompts ($K$) of 16 and a weight of empirical concept ($\eta$) of 3.
For our evaluation, we generate one image for each of these 95 adversarial prompts, using the fixed seeds provided by RAB for reproducibility. The resulting images are subsequently evaluated using NudeNet to calculate the ASR.
\item \textbf{CCE}~\cite{pham2023circumventing}: This white-box attack employs Textual Inversion to optimize a new embedding vector ($v_a^*$) that serves as a proxy for the erased concept.
For the Nudity task, to ensure a fair comparison with other baselines, we use the same 142 images from our UDA setup as the training set to optimize $v_a^*$ on each sanitized model. For evaluation, we then prepend the learned $v_a^*$ to the 142 corresponding prompts.
For the Artistic Style (Van Gogh) task, we follow the original CCE protocol~\cite{pham2023circumventing}, using 6 images generated from ``A painting in the style of Van Gogh'' as the training set. We then evaluate by generating 50 images using the prompt ``a painting in the style of $v_a^*$''.
For the Object (tench) task, we again adhere to the original CCE protocol, using 30 images of the object as the training set. Evaluation is then performed by generating 50 images using the prompt ``A photo of $v_a^*$'', aligning with the evaluation scale of other baselines.
All inversion optimizations are trained for 1,000 steps with a learning rate of 0.005.
\end{itemize}

\section{Ablation Study}
\label{sec:appx_ablation}

\begin{figure}[t]
    \centering
    \includegraphics[width=\linewidth]{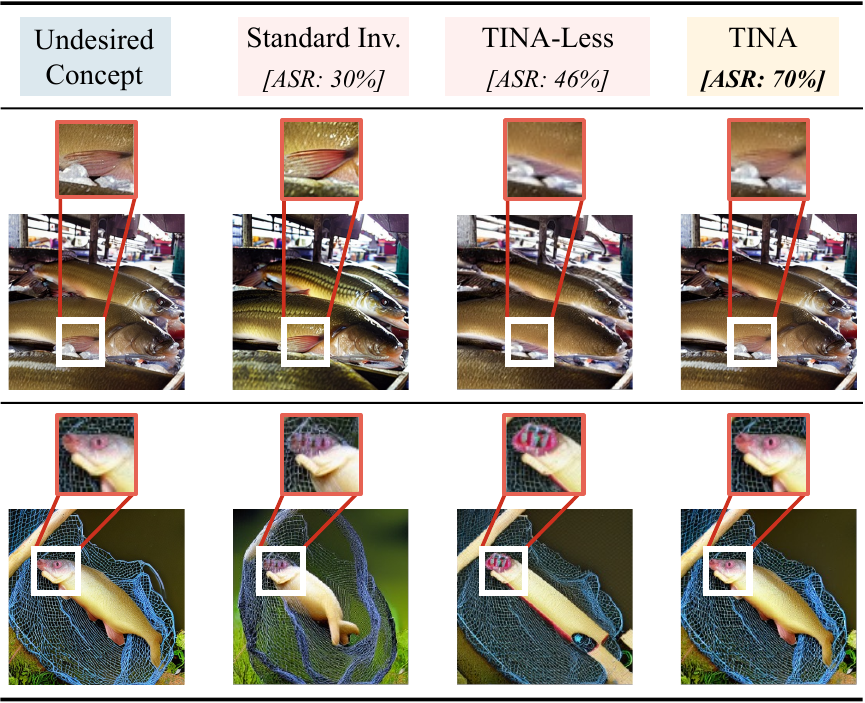}
    \caption{Ablation study of attack results on the ESD method for tench object erasure. From left to right: target concept, \texttt{Standard Inv.} (standard text-guided inversion), \texttt{TINA-Less} (with insufficient optimization), and our full TINA method.}
    \label{fig:ablation}
\end{figure}

To validate the necessity of our optimization procedure, we conduct an ablation study comparing our full TINA framework against two baselines (Figure~\ref{fig:ablation}): (1) a standard, text-guided DDIM inversion (\texttt{Standard Inv.}) and (2) a variant of TINA with insufficient optimization steps (\texttt{TINA-Less}).
The results clearly demonstrate the importance of sufficient optimization.
The text-guided \texttt{Standard Inv.} fails (30\% ASR), as the erasure method actively counteracts the textual prompt.
\texttt{TINA-Less}, which lacks sufficient optimization iterations to correct cumulative approximation errors, also performs poorly (46\% ASR), producing distorted reconstructions.
In contrast, our full TINA, which applies adequate optimization iterations to refine each inversion step, achieves high-fidelity reconstruction and a 70\% ASR.
This 24-point ASR improvement over \texttt{TINA-Less} demonstrates that a sufficient number of optimization steps is critical for accurately identifying latent generative trajectories, confirming the necessity of our iterative refinement procedure.

\begin{figure}[t]
    \centering
    \includegraphics[width=\linewidth]{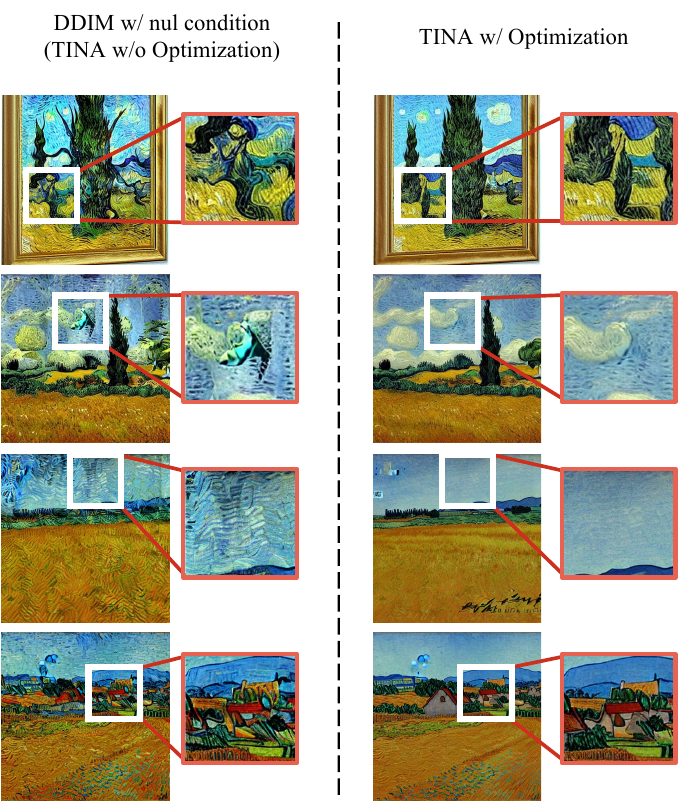}
    \caption{Visual comparison between naive null-text DDIM inversion and our optimized TINA inversion on the Van Gogh style erasure task. While the naive null-text DDIM inversion can roughly recover the global composition, it often fails to preserve fine-grained stylistic details, such as brushstroke textures and local structures. In contrast, TINA produces reconstructions with noticeably richer and more faithful visual details.}
    \label{fig:null_ddim_tina}
\end{figure}

Figure~\ref{fig:null_ddim_tina} provides a complementary visual perspective on the role of optimization. When the text condition is directly set to null, the resulting inversion can still recover the coarse layout of the target image, but the reconstructed results often exhibit blurred local structures and weakened style-specific textures. This behavior indicates that the naive null-text DDIM inversion does not reliably trace a sufficiently precise generative trajectory for faithful concept recovery. By contrast, the optimization procedure in TINA yields reconstructions with sharper local patterns and more coherent stylistic cues, showing that its benefit lies not only in recovering the concept at a coarse semantic level, but also in preserving the fine-grained visual details that define the target concept.

\section{Comparison with DDIM-Based Reconstruction Methods}
\label{sec:ddim_comparison}

To further contextualize the performance of TINA, we compare it against a recent state-of-the-art DDIM-based reconstruction method, EasyInv~\cite{zhang2025easyinv}, which has been shown to outperform vanilla DDIM Inversion~\cite{song2020denoising}, Null-Text Inversion~\cite{mokady2023null}, Negative-Prompt Inversion~\cite{miyake2025negative}, PnP Inversion~\cite{ju2023pnp}, and others.
To ensure a fair comparison, we configure EasyInv to operate under the same text-free condition as TINA.
As shown in Table~\ref{tab:easyinv}, our TINA significantly outperforms EasyInv across all five unlearning defenses on the tench object erasure task.
This substantial performance gap demonstrates that general-purpose DDIM-based reconstruction methods, even state-of-the-art ones, are insufficient for the concept recovery task, further validating the necessity of our tailored optimization-based inversion approach.

\begin{table}[t]
  \centering
    \begin{tabular}{lcc}
    \toprule
          & \textbf{EasyInv}~\cite{zhang2025easyinv} & \textbf{TINA} \\
    \midrule
    \textbf{ESD}~\cite{gandikota2023erasing} & 24.0  & \textbf{70.0} \\
    \textbf{EraseDiff}~\cite{wu2024erasediff} & 26.0  & \textbf{68.0} \\
    \textbf{SalUn}~\cite{fan2023salun} & 30.0  & \textbf{72.0} \\
    \textbf{Scissorhands}~\cite{wu2024scissorhands} & 34.0  & \textbf{78.0} \\
    \textbf{STEREO}~\cite{srivatsan2025stereo} & 24.0  & \textbf{72.0} \\
    \bottomrule
    \end{tabular}%
  \caption{Attack Success Rates (ASR, in \%) comparison between EasyInv and our TINA on the tench object erasure task. Both methods operate under a text-free condition. \textbf{Bold} indicates the best-performing method.}
  \label{tab:easyinv}%
\end{table}%

\section{Generalization to DiT-Based Architectures}
\label{sec:dit_generalization}

While the main experiments focus on UNet-based diffusion models (\eg, Stable Diffusion v1.4), an important question is whether TINA generalizes to fundamentally different model architectures. To investigate this, we evaluate TINA on a Diffusion Transformer (DiT)~\cite{peebles2023scalable}-based text-to-image model, specifically \texttt{PixArt-XL-2-512x512}~\cite{chen2023pixart}, which replaces the UNet backbone with a transformer architecture.

Since \texttt{PixArt-XL-2-512x512} cannot generate the ``Tench'' concept used in our main experiments, we instead conduct this evaluation on the ``Parachute'' concept.
Additionally, as there are no publicly available erased versions of this DiT model, we first apply the ESD~\cite{gandikota2023erasing} method to erase the parachute concept from the model, and then perform our TINA attack on the resulting erased model.

\begin{figure}[t]
    \centering
    \includegraphics[width=\linewidth]{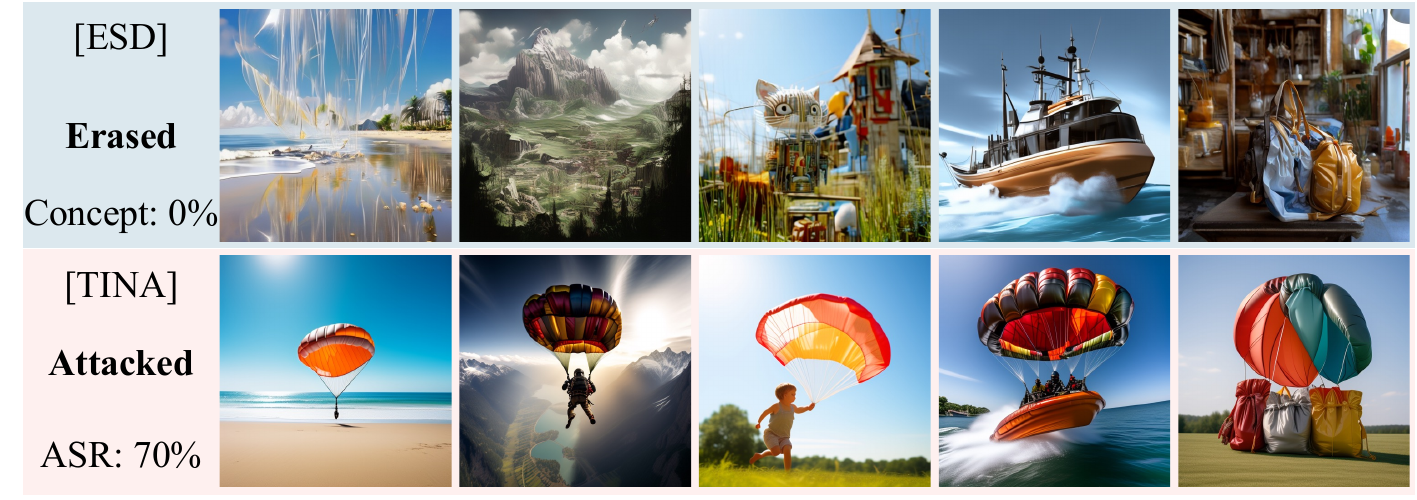}
    \caption{Generalization of TINA to a DiT-based architecture (\texttt{PixArt-XL-2-512x512}). We erase the ``Parachute'' concept using ESD and then apply TINA. While the erased model fails to generate parachutes, TINA successfully recovers the concept, demonstrating architecture-agnostic generalizability.}
    \label{fig:dit_generalization}
\end{figure}

As shown in Figure~\ref{fig:dit_generalization}, the erased DiT model is no longer able to generate the parachute concept under normal sampling conditions.
However, our TINA method successfully recovers the erased concept, producing recognizable parachute images from the erased model.
This result demonstrates that the vulnerability exposed by TINA is not limited to UNet-based architectures, but extends to DiT-based models as well, confirming the broad generalizability of our attack.

\section{Failure Cases}
\label{sec:failure_cases}

While TINA consistently achieves the highest ASR across all evaluated defenses, its performance is not uniform.
As reported in Table~\ref{tab:style}, TINA's ASR against STEREO~\cite{srivatsan2025stereo} on the artistic style erasure task is $44.0\%$, which, despite being the best among all evaluated attacks, is notably lower than its $70.0\%$--$80.0\%$ ASR against other defenses.
This indicates that adversarially robust erasure methods can partially resist our text-free inversion attack.

\begin{figure}[t]
    \centering
    \includegraphics[width=\linewidth]{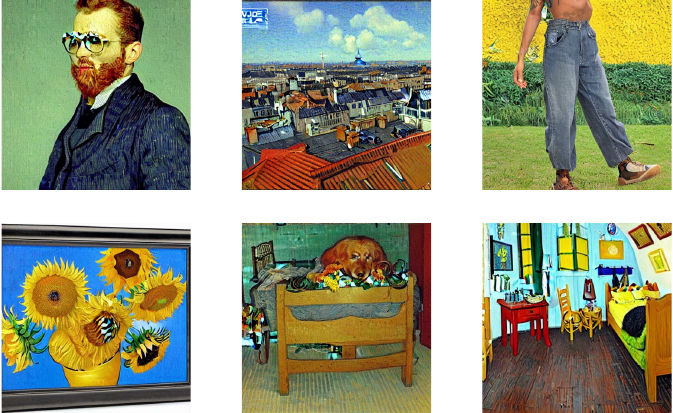}
    \caption{Failure cases of TINA on the STEREO style erasure (Van Gogh) task. While TINA preserves the spatial composition of the target images, the reconstructed results fail to recover the distinctive artistic style, indicating that the adversarial training procedure of STEREO partially disrupts the style-specific visual knowledge.}
    \label{fig:failure_case}
\end{figure}

To provide a concrete illustration, Figure~\ref{fig:failure_case} presents several failure cases drawn from the STEREO style erasure setting.
In these examples, images reconstructed by TINA preserve the overall spatial structure and composition of the target, yet fail to recover the distinctive characteristics of Van Gogh style, such as the expressive brushwork, bold color palette, and textural impasto.
We attribute this partial failure to two factors.
First, the adversarial training procedure of STEREO not only fortifies the text-image mapping against adversarial prompts but may also perturb the internal visual representations, making the stylistic knowledge harder to reactivate via a text-free trajectory.
Second, unlike concrete objects or explicit content, artistic style is a high-level, distributed visual attribute whose encoding in the parameter space is inherently more diffuse and thus more susceptible to robust erasure.
This also explains the discrepancy with the object erasure task, where TINA still achieves $72.0\%$ ASR against the same STEREO defense (Table~\ref{tab:object}).

Nevertheless, these failure cases do not undermine our central thesis.
Even in the most challenging setting, a $44.0\%$ ASR of TINA demonstrates that a substantial portion of the visual knowledge persists within the erased model.
Moreover, all text-centric baselines are entirely neutralized by STEREO on style erasure ($0.0\%$ for both P4D and UDA), whereas TINA still succeeds in a significant fraction of cases.
These results suggest that while adversarially robust methods like STEREO represent a meaningful step forward, they still fall short of fully eliminating visual knowledge, reinforcing the need for erasure paradigms that operate directly on internal visual representations.

\begin{figure*}[ht]
    \centering
    \includegraphics[width=\linewidth]{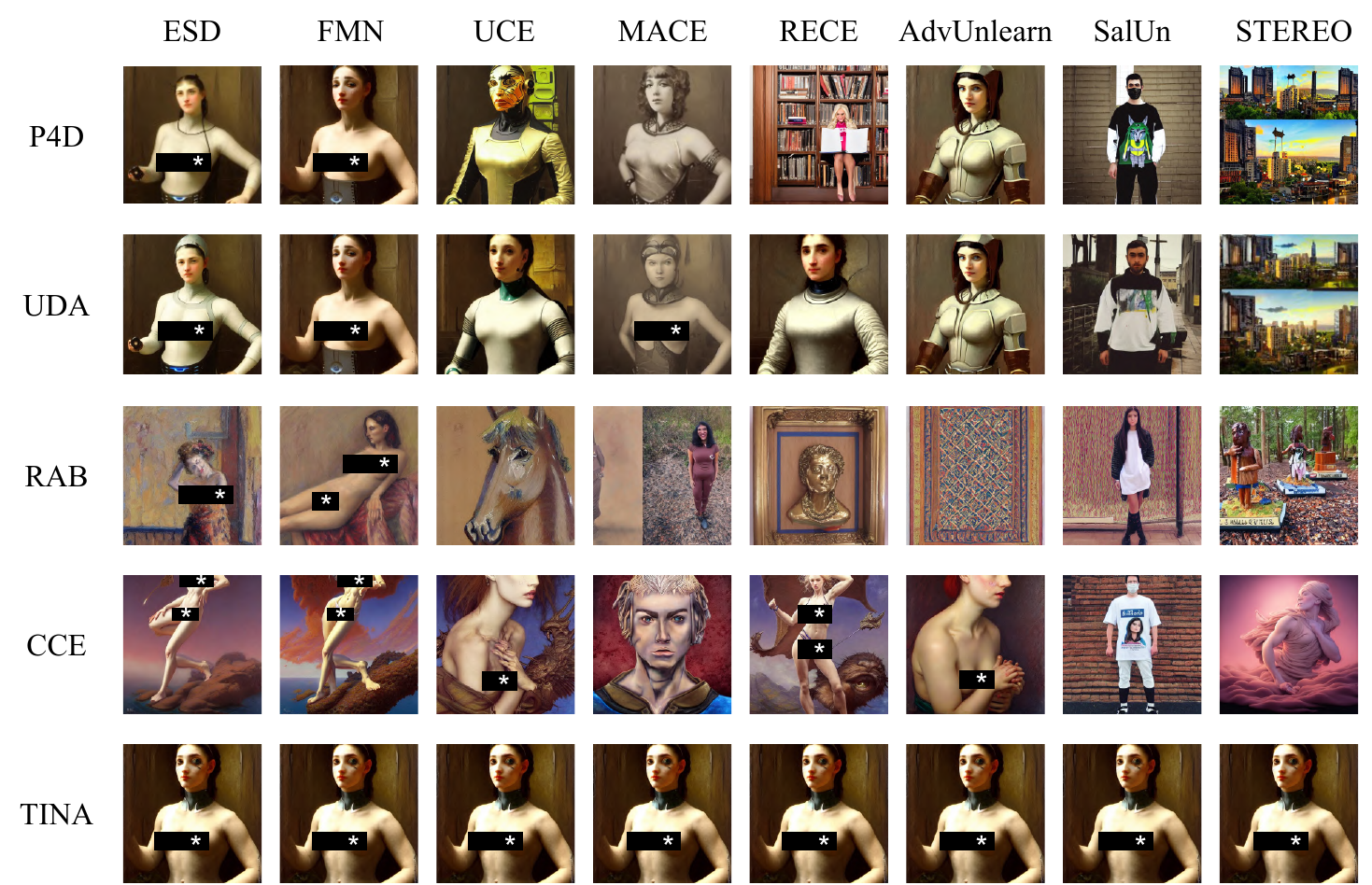}
    \caption{Qualitative comparison of attack methods on the nudity erasure task. Each row corresponds to an attack method, and each column corresponds to an unlearning defense. Sensitive content is redacted.}
    \label{fig:more_nudity}
\end{figure*}

\section{More Visual Results}
\label{sec:more_visual_results}

\begin{figure*}[ht]
    \centering
    \includegraphics[width=\linewidth]{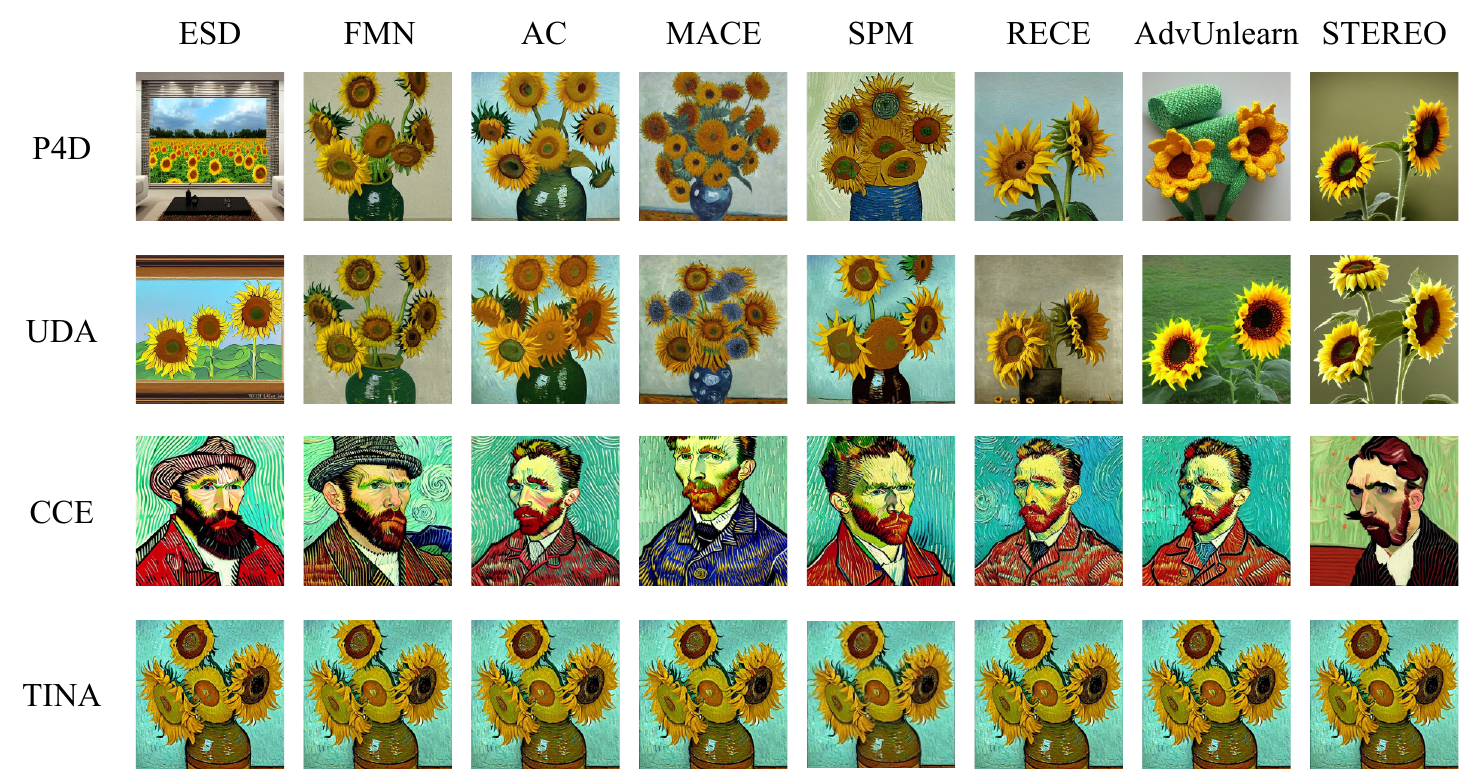}
    \caption{Qualitative comparison of attack methods on the ``Van Gogh'' style erasure task. Each row represents an attack, and each column represents a specific unlearning defense.}
    \label{fig:more_vangogh}
\end{figure*}

We provide expanded qualitative comparisons in Figure~\ref{fig:more_nudity} and Figure~\ref{fig:more_vangogh} to further demonstrate the fundamental vulnerability exposed by TINA.

Figure~\ref{fig:more_nudity} presents a comprehensive matrix of attack results for the nudity erasure task. These visualizations visually substantiate our quantitative findings. A clear trend emerges for text-centric baselines (P4D, UDA, RAB, and CCE): while they show varying degrees of success against simpler defenses, their efficacy diminishes significantly against more advanced, adversarially-aware methods. Specifically, against robust defenses such as AdvUnlearn, SalUn, and STEREO, these attacks are almost entirely mitigated, producing neutral or unrelated imagery. In stark contrast, the bottom row shows that our TINA consistently bypasses all defenses, successfully uncovering the generative process to regenerate the forbidden content.

Figure~\ref{fig:more_vangogh} details the results for the Van Gogh style erasure. This figure not only reinforces the failure of text-based attacks (P4D, UDA) against robust defenses (AdvUnlearn, STEREO) but also reveals a critical flaw in embedding-space attacks. As shown in the third row, the CCE attack conflates the concept of ``Van Gogh's style'' with ``Van Gogh's likeness.'' Consequently, it predominantly generates portraits of Van Gogh himself, rather than applying the distinct artistic style to a general concept. TINA, however, does not suffer from this ambiguity. Operating independently of the text-conditioning path, TINA (bottom row) successfully isolates and reactivates the underlying generative trajectory for the style itself, proving that this visual knowledge persists even after robust erasure.

\end{document}